# OVPD: A Virtual-Physical Fusion Testing Dataset of OnSite Autonomous Driving Challenge

Yuhang. Zhang[1], Jiarui. Zhang[1], Bowen. Jian[1], Xin. Zhou[1], Zhichao. Lv[1], Peng. Hang[1], Rongjie. Yu[1], Ye. Tian[1] and Jian. Sun[1,✉]

**ABSTRACT:** The rapid iteration of autonomous driving algorithms has created a growing demand for high-fidelity, replayable, and diagnosable testing data. However, many public datasets lack real vehicle dynamics feedback and closed-loop interaction with surrounding traffic and road infrastructure, limiting their ability to reflect deployment readiness. To address this gap, we present OVPD (OnSite Virtual–Physical Dataset), a virtual–physical fusion testing dataset released from the 2025 OnSite Autonomous Driving Challenge. Centered on real-vehicle-in-the-loop testing, OVPD integrates virtual background traffic with vehicle–infrastructure perception to build controllable and interactive closed-loop test environments on a proving ground. The dataset contains 20 testing clips from 20 teams over a scenario chain of 15 atomic scenarios, totaling nearly 3 hours of multi-modal data, including vehicle trajectories and states, control commands, and digital-twin-rendered surround-view observations. OVPD supports long-tail planning and decision-making validation, open-loop or platform-enabled closed-loop evaluation, and comprehensive assessment across safety, efficiency, comfort, rule compliance, and traffic impact, providing actionable evidence for failure diagnosis and iterative improvement. The dataset is available via: https://huggingface.co/datasets/Yuhang253820/Onsite_OPVD

## 1 Introduction

Autonomous driving is evolving from driver assistance toward L3+ autonomy, which imposes stricter requirements on complex, interaction-intensive, and traffic-rule-constrained urban driving (SAE International, 2021). However, mainstream algorithm development still relies heavily on long-horizon road testing and large-scale real-world data accumulation: fleets collect driving logs, from which ``valuable'' segments are mined for model training or replay-based validation (Lou et al., 2022). While this paradigm continuously increases data volume, it faces two fundamental bottlenecks. First, long-tail, high-risk scenarios occur extremely rarely in naturalistic driving, making systematic coverage difficult with passive collection alone (Sun et al., 2021). Second, evaluation outcomes are often weakly diagnostic, failing to locate missing capabilities under certain scenario, which in turn slows down issue localization and iteration. As the marginal returns of scaling data begin to saturate, simply collecting ``more data'' no longer yields proportional gains in test coverage, creating an urgent need for training or testing data that are more targeted and controllable (Cui et al., 2025; Hafner et al., 2025).

Meanwhile, using simulated scenario data for training still suffers from a pronounced sim-to-real gap, even with physics-enabled physics-enabled simulators, particularly in vehicle dynamics, execution latency, and control stability (Daza et al., 2023; Stocco et al., 2022; Tristano et al., 2024). In complex interactive scenarios, control difficulty grows exponentially compared to simpler cases, whereas debugging directly on public roads is costly and introduces substantial safety risks (National Transportation Safety Board, 2019; Sadigh et al., 2016).

Existing evaluation practices further amplify the above gaps. When assessing an autonomous driving algorithm, one line of practice uses a route or a composition of scenarios as the evaluation unit and computes a driving score in interactive closed-loop replay (Jia et al., 2024); another line adopts an open-loop setting and measures planning fidelity via regression error between the planned trajectory and logged ground truth (e.g., L2 error) (Karnchanachari et al., 2024). Conventional metric suites mostly emphasize ego-centric planning accuracy, safety, efficiency, and comfort, but they are less capable of capturing deployment-critical aspects such as compliance with traffic rules and the algorithm's impact on surrounding traffic flow (Li et al., 2024; Zhai et al., 2023). As a result, they may encourage strategies that appear ``safe'' yet are socially or operationally unreasonable---for example, rushing for gaps, coercive merges, or overly conservative behaviors that hinder overall traffic efficiency (Jia et al. 2024; Yu et al. 2021), More over, route-level aggregation makes it difficult to localize capability deficits at the scenario level (Ericsson, 2000).

[1] College of Transportation, Tongji University, Shanghai 201804, China.
✉ Corresponding author. E-mail: {zhangyvhang, zjr0915, jianbowen, zhouxin517, lvzhichao, hangpeng, yurongjie, tianye, sunjian}@tongji.edu.cn

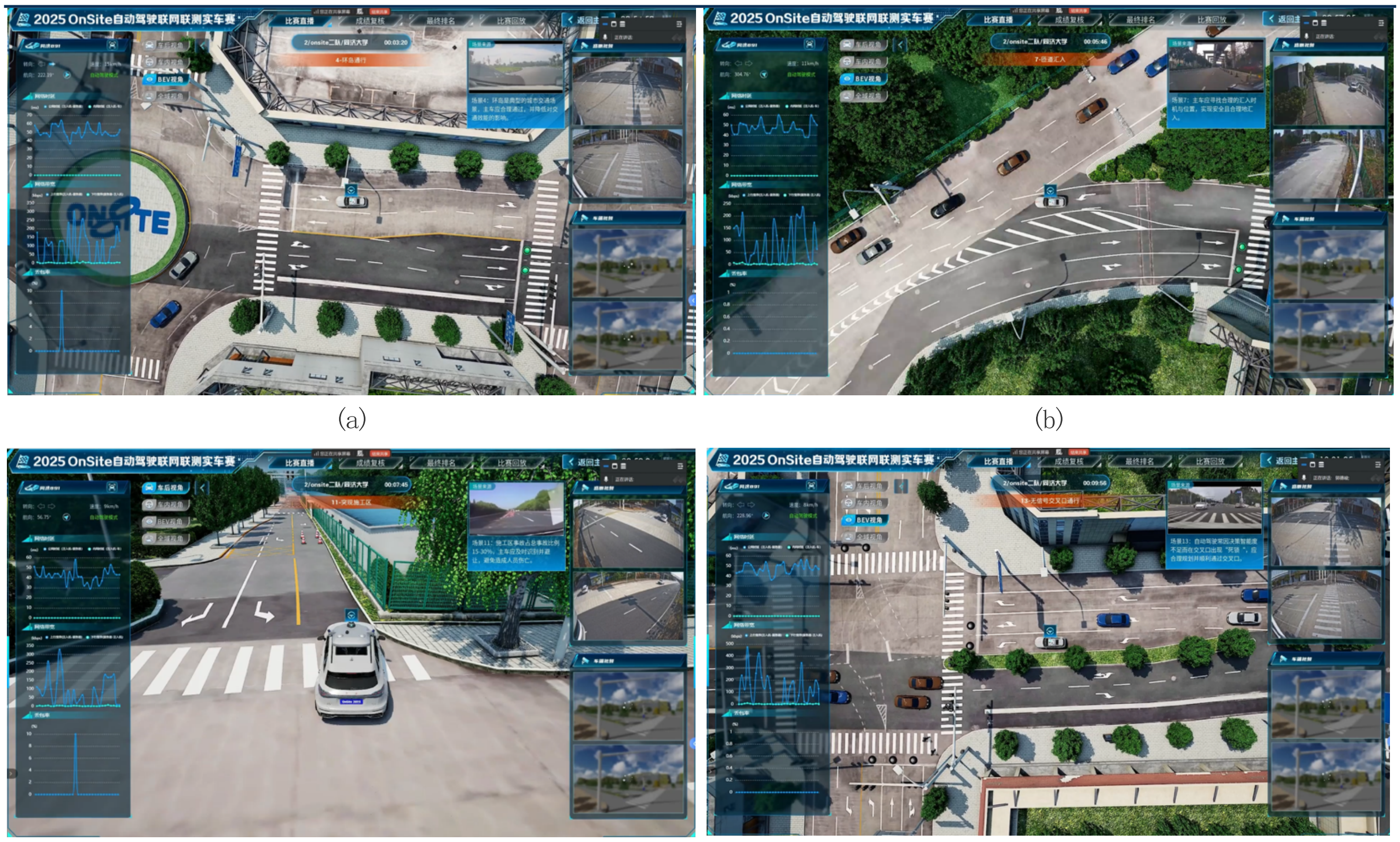

(a) (b) (c) (d)

**Fig. 1.** A few exemplary visualizations of the OVPD dataset: (a) a roundabout scenario, highlighting yielding and interaction with circulating traffic; (b) a merge scenario, focusing on gap selection and coordination with surrounding vehicles; (c) an obstacle-avoidance scenario in a construction zone; and (d) an unsignalized intersection scenario, emphasizing right-of-way judgment and conflict negotiation.

To this end, we release **OVPD**, a virtual–physical fusion testing dataset derived from the 2025 Onsite Autonomous Driving Challenge (Research Team for Traffic OPerations and Simulation, 2025). The Onsite real-vehicle autonomous driving challenge was jointly organized by Tongji University and the Ministry of Transport of China, under the guidance of SAE China and the National Natural Science Foundation of China (NSFC). Targeting the multi-dimensional needs of autonomous driving product certification, market access, and operational supervision, the competition aims to explore a cross-regional and standardized joint evaluation framework for autonomous driving systems. Prior to the on-site vehicle competition, an online algorithm challenge was conducted to screen participants, from which 22 outstanding teams were selected for the real-vehicle stage.

The competition adopts a virtual–physical fusion evaluation paradigm that combines simulation-based testing with proving-ground testing, and is built upon the **VP-AutoTest** platform (Cui et al., 2025). Centered on real-vehicle-in-the-loop execution, it integrates virtual background traffic flow with vehicle–roadside perception to construct an interactive testing environment in a closed proving ground.

The OVPD dataset contains test data from 20 participating teams collected on a scenario chain composed of 15 atomic scenarios, nearly 3 hours. OVPD fully open-sources vehicle trajectories and states, control commands, as well as virtual camera views. It provides a unified data foundation for the training, testing, and diagnostic analysis of autonomous driving algorithms.

OVPD provides data support to address the above challenges through three core contributions:

**1) Capability-oriented scenario generation pipeline:** The scenarios in OVPD are generated via a capability-driven scenario construction pipeline. We establish a hierarchical taxonomy along three dimensions—basic driving functions, road facility types, and core driving capabilities—so as to build a closed-loop scenario library that is capability-targeted, difficulty-adjustable, and controllable, enabling precise diagnosis of algorithmic weaknesses.

**2) Multi-dimensional metrics for diagnosis:** Beyond foundational dimensions such as safety, efficiency, and comfort, we introduce metrics for traffic-rule compliance and traffic coordination, providing a more comprehensive characterization of deployability under real urban traffic constraints.

**3) Real vehicle dynamics constraints in the data:** OVPD is collected in a virtual--physical fusion testing environment centered on real-vehicle-in-the-loop execution. The ego vehicle interacts with virtual background traffic while producing real state, trajectory, and control data, enabling the dataset to capture realistic dynamic constraints for model training and validation.

## 2 Related Works

### *2.1. Autonomous Driving Datasets*

In recent years, the field has accumulated a large number of datasets with increasing scale and broader scenario coverage. Open-loop datasets (Caesar et al., 2020; Sun et al., 2020; Wang et al., 2019) represented by nuScenes typically evaluate planned trajectories through regression errors against logged ground truth (e.g., L2 error), offering advantages in implementation simplicity and computational efficiency. Large-scale datasets such as Waymo (Ettinger et al., 2021; Xu et al., 2025) further provide extensive scenario diversity and rich multi-agent motion data, forming an important foundation for perception, prediction, and planning research.

However, despite their large scale and broad coverage, existing autonomous driving datasets are still predominantly designed for modeling or general benchmark evaluation. There remains a lack of targeted testing datasets explicitly oriented toward capability diagnosis and deployment-critical assessment. In particular, test resources for capabilities such as emergency collision avoidance, traffic-rule compliance, and traffic coordination remain relatively limited, even though these aspects are among the most critical concerns for real-world autonomous driving deployment. Existing open-loop settings correlate only weakly with safety risks, traffic efficiency, and socially compliant behaviors that emerge in closed-loop interaction (Jia et al., 2024; Li et al., 2024), while many closed-loop benchmarks still organize evaluation primarily at the route or long-horizon replay level, making it difficult to conduct controllable, difficulty-adjustable, and capability-targeted testing (Cui et al., 2025; Guo et al., 2025; Yang et al., 2024; Zhou et al., 2025).

NuPlan (Karnchanachari et al., 2024), as a widely used planning benchmark, provides datasets, maps, and a simulation evaluation stack, yielding evaluations closer to real driving. Bench2Drive (Jia et al., 2024) further targets end-to-end closed-loop evaluation and introduces comfort- and efficiency-related metrics to characterize driving experience. Nevertheless, such benchmarks still provide insufficient coverage of deployment-critical dimensions such as emergency risk handling, traffic-rule compliance, and interaction coordination (Jiang et al., 2025; Kurenkov et al., 2024; United Nations Economic Commission for Europe, 2025). As a result, methods optimized under existing benchmarks may achieve favorable scores while still exhibiting undesirable real-world behaviors, such as aggressive gap-taking, rule-violating maneuvers, or blocking surrounding traffic (Crosato et al., 2023; Zhou et al., 2024).

To bridge these gaps, OVPD makes new attempts in both data organization and metric outputs (OnSite Autonomous Driving Public Testing Service Platform, n.d.). OVPD adopts scenario-centric clip units and leverages a capability-oriented scenario construction and grading strategy to improve long-tail coverage for training, and to enhance controllability and diagnostic granularity for evaluation. In addition to conventional dimensions, OVPD introduces traffic-rule compliance and traffic coordination metrics to better reflect deployability on urban roads.

### *2.2. Scenario Generation and Simulation Tools*

Scenario generation and simulation tools provide an essential foundation for the development and validation of autonomous driving algorithms. Traditional approaches mainly rely on rule-driven or physics-engine-based simulation platforms (Dosovitskiy et al., 2017; Fan et al., 2018; Zhang et al., 2025). These platforms typically construct interactive environments based on predefined maps, traffic participant behavior models, and vehicle dynamics models, enabling large-scale scenario replay, algorithm debugging, and closed-loop testing at relatively low cost. However, conventional simulation tools still suffer from limitations in rendering realism, behavioral diversity of traffic participants, and realistic vehicle dynamic feedback, especially in highly interactive scenarios where a clear gap remains between simulation and real-world deployment (Garcia Daza et al., 2023; Stocco et al., 2023).

In recent years, with the rapid development of generative modeling and neural scene representation techniques, autonomous driving simulation has been evolving toward higher fidelity. Environment generation methods based on world models (Gao et al., 2024; Zhao et al., 2025), as well as scene reconstruction techniques combined with 3D Gaussian Splatting (3DGS) (Kerbl et al., 2023; Yan et al., 2024), have shown greater potential in visual realism, dynamic interaction representation, and diversified data generation.

Nevertheless, most existing scenario generation and simulation tools still focus primarily on virtual environment construction itself, and are therefore more suitable for offline training, purely simulated validation, or visual reconstruction. Their support remains limited for deployment-critical requirements, such as realistic vehicle dynamics constraints, real-time control loops, and closed-loop interaction between virtual and physical traffic participants. In contrast, the VP-AutoTest platform (Cui et al., 2025) underlying OVPD constructs a virtual–physical fusion environment based on digital twin technology. By integrating high-definition maps, roadside perception, virtual background traffic flow, and real vehicle control, it enables real-time closed-loop interaction between virtual and physical traffic participants. The platform can not only generate and collect high-fidelity data for training and testing, but also directly serve as a unified environment for pre-training validation and closed-loop evaluation.

## 3 Construction

In this section, we introduce the pipeline of dataset construction, including scenario designation, evaluation metrics, visual-physical elements, data processing, etc.

### *3.1. Testing Scenarios*

The OVPD dataset is collected from the on-vehicle test records of the 2025 Onsite Autonomous Driving Challenge (Research Team for Traffic OPerations and Simulation, 2025). The dataset is organized into replay clips, with 20 clips in total, each corresponding to one complete test run from a participating team. Each clip lasts about 10 minutes (approximately 6000 frames) and all modalities are aligned using a unified global timestamp.

Each clip contains a scenario chain formed by concatenating 15 representative test scenarios, which are designed by an expert panel and illustrated in Fig. 2 and Table 1. These scenarios are centered on representative challenging events and common capability bottlenecks in urban autonomous driving. Under constraints imposed by road infrastructure, traffic rules, and multi-agent interaction patterns, they induce dense critical decisions and interactions, thereby providing planning and decision-making data with strong diagnostic value.

All scenarios are organized according to the road topology of the test field: each atomic scenario is defined with an explicit trigger point and termination point, and scenarios are connected end-to-end on the lane-level topology, thereby forming a complete route within a unified test map.

In this paper, we summarize the 15 baseline scenarios using a ``short name + one-sentence definition" format. The scenarios can be grouped as follows (note that the order here is independent of the execution order in the scenario chain).

**Table 1.** Summary of Typical Autonomous Driving Test Scenarios

| | Scenario Name and Description |
|---|---|
| **(A) Emergency-Response Scenarios** | |
| A1 | **Pedestrian pop-out crossing:** Pedestrians suddenly enter the ego lane from roadside under limited visibility. The ego must rapidly infer intent and perform emergency braking/yielding or lateral evasion when necessary. |
| A2 | **Suddenly exposed stalled vehicle:** A stopped broken-down vehicle is revealed after the lead vehicle changes lanes. The ego must recognize the stationary obstacle and mitigate risk within a short distance. |
| A3 | **Red-light-running emergency vehicle:** During a left turn at a T-junction, an occluded ambulance from the left approach runs a red light. The ego must react promptly to avoid collision. |
| A4 | **Lead vehicle hard braking:** The lead vehicle brakes sharply during car-following. The ego must choose braking/brake-and-lane-change/decelerated following conditioned on headway and surrounding traffic. |
| A5 | **Aggressive e-scooter cut-in:** An e-scooter rapidly cuts in from the non-motorized lane. The ego must predict near-term intent within a short reaction window and avoid collision. |
| A6 | **Fast background-vehicle cut-in:** A side vehicle accelerates and forces a merge into the ego lane. The ego must detect cut-in intent, anticipate conflict, and execute a reasonable evasive strategy. |
| A7 | **Work-zone obstacle avoidance:** An unannotated work-zone occupation appears ahead. The ego must detect it, locally replan, and pass safely via deceleration and/or lane change. |
| **(B) Traffic-Efficiency Scenarios** | |
| B1 | **Roundabout traversal**: The ego yields to circulating vehicles, enters at an appropriate gap, maintains a reasonable speed inside the roundabout, and plans an efficient exit while avoiding conflicts. |
| B2 | **Ramp merging:** With high-speed mainline flow and fast-approaching ramp vehicles, the ego must perform multi-agent prediction and timely longitudinal/lateral decisions to merge safely without significantly disturbing traffic flow. |
| B3 | **Non-motorized spillover:** A blocked non-motorized stream suddenly spills into the motorized lane. The ego must anticipate lateral motion and respond (braking/deceleration/lane change) without creating a moving bottleneck. |
| B4 | **Unsignalized intersection negotiation:** The ego plans a left turn while vehicles from other directions and pedestrians create repeated right-of-way interactions. The ego must pass safely without being overly conservative and blocking traffic. |
| **(C) Rule-Compliance Scenarios** | |
| C1 | **Consecutive right turns with blockage:** A failure vehicle blocks the right-turn lane. The ego must proceed under constraints, evaluating necessary rule deviation under safety-first when compliance conflicts with task reachability (criteria defined in the metric section) |
| C2 | **Queue spillback at an intersection:** After the green phase, downstream congestion leaves limited space; crossing the stop line risks illegal stopping inside the intersection and blockage. The ego should wait until traffic clears. |
| C3 | **Complex signal-phase composition:** The signal switches from right-turn prohibited to right-turn permitted. The ego must interpret signal constraints correctly and yield to pedestrians during the turn. |
| C4 | **Autonomous parking under restrictions:** Roadside constraints (bus stop, solid yellow line, hatched markings) prohibit stopping. The ego must recognize markings and select a legal parking area to complete the task. |

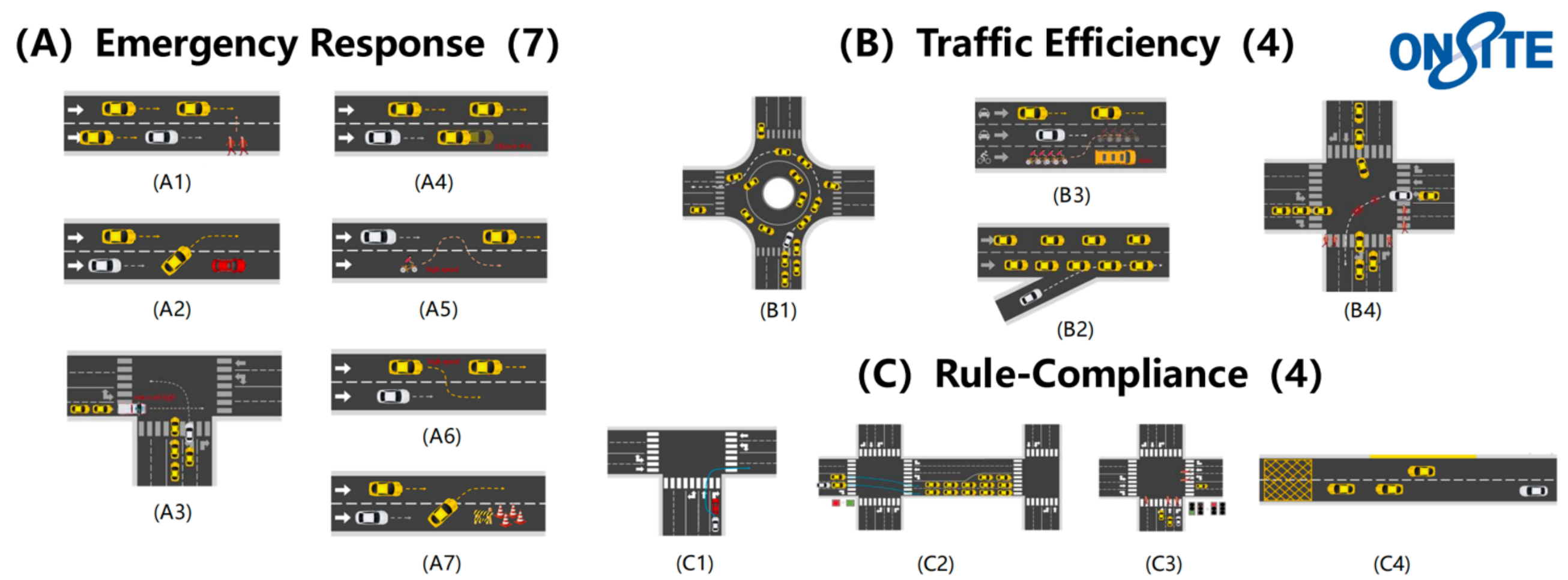

**Fig. 2.** Illustration of the 15 atomic test scenarios included in OVPD. All scenarios are organized into three capability-oriented categories: Emergency Response (A1–A7), Traffic Efficiency (B1–B4), and Rule Compliance (C1–C4). Together, these scenarios form the testing basis of OVPD for deployment-critical capabilities in urban driving.

### 3.2. Virtual-Physical Data Recording

Enabled by the Onsite on-vehicle challenge and the virtual--physical fusion testing platform (VP-Test/VP-AutoTest), the ego-vehicle portion of OPVD is collected from real executions of deployed autonomous driving stacks. As a result, the dataset captures not only trajectory-level behavioral outcomes but also the feedback and constraints introduced by real vehicle dynamics and the actuation/execution pipeline. This enables analysis of controlling stability and robustness under real-time closed-loop constraints, such as variations in speed and acceleration, steering response characteristics, and the evolution of vehicle pose along the tracking process.

At the data generation and replay stage, the testing procedure is orchestrated by the Onsite Open Platform. Background traffic flows are generated by TESSNG and, together with the ego-vehicle perception information, are fed into the ego domain controller to enable interactive closed-loop execution. The dataset further provides platform-rendered surround-view and BEV views at 10~Hz, supporting joint analysis and modeling across *trajectory--control--perception representations*. In addition, each clip is accompanied by aggregated scores along five dimensions, which characterize a team's overall performance over the complete scenario chain.

The virtual--physical communication and data transmission on the platform side are implemented based on the VP-AutoTest (Cui et al., 2025) system, and are integrated with the onboard perception-fusion capability to form a complete testing toolchain that supports Onsite scenario execution and OPVD data logging. The ego vehicle's control commands and motion states are recorded at 10~Hz: high-level planning/decision modules output control commands, which are executed through the onboard controller and the drive-by-wire chassis in a low-level closed loop.

### 3.3. Comprehensive Evaluation Metrics

In addition to the scenario test logs, the OVPD dataset provides score outputs associated with each test record. For each clip, we release the evaluation metric results produced in the Onsite Autonomous Driving Challenge, together with the computation protocols of all metrics, to support offline replay and analysis by researchers.

Existing planning datasets and metrics offer limited relevance to travel efficiency and socially compliant driving behaviors. To better characterize the deployability of planning/decision-making algorithms in complex urban interactions, OVPD adopts five complementary dimensions to score performance within each individual scenario: *Efficiency, Safety, Comfort, Rule Compliance, and Traffic Coordination.*

Moreover, OVPD fixes the scoring granularity at the *single-scenario* level and aggregates scores thereafter, bringing two key benefits: (i) **strong binding between scores and scenario semantics**---a scenario-level score directly reflects the capability requirements of that scenario, making performance variations easier to map to specific interaction patterns and rule constraints for bottleneck diagnosis; and (ii) **more stable cross-method comparison**---it mitigates the disproportionate impact of sporadic violations along long routes, yielding comparisons that are more robust and interpretable (Jia et al., 2024).

The detailed metric definitions are given as follows:

Let a scenario chain consist of $N$ atomic scenarios, and denote the overall score of the $i$-th scenario by $S_i$. We compute the scenario-chain score and the per-dimension scores via **uniform averaging over scenarios**, i.e.

**Overall Driving Score:** $\mathrm{Score}_{\mathrm{overall}} = \frac{1}{N}\sum_{i=1}^{N} S_i$

**Per-dimension scores (five dimensions):** for each dimension, we apply the same scenario-wise averaging to obtain per-dimension leaderboards, which are further used to generate capability diagnostic reports.

Detailed calculation methods are following:

1) *Single-Scenario Scoring Formula*: The overall score of a single scenario uses the efficiency term as a baseline and combines the other four dimensions through a weighted coefficient. The weights are determined by expert review and follow the common principle adopted in related benchmarks---*safety first, comfort next, with rules and interaction as complementary factors*. Specifically, the single-scenario score is defined as

$$S = G_{\mathrm{eff}}^{0.5} \times \left(0.5\, G_{\mathrm{safety}} + 0.3\, G_{\mathrm{comfort}} + 0.1\, G_{\mathrm{rule}} + 0.1\, G_{\mathrm{coord}}\right) \quad (1)$$

where,

$G_{\mathrm{eff}} \in [0,100]$ is the efficiency score (with 100 as the maximum);

$G_{\mathrm{safety}}, G_{\mathrm{comfort}}, G_{\mathrm{rule}}, G_{\mathrm{coord}} \in [0,10]$ are the other four scores (with 10 as the maximum).

We apply a square-root transform (power 0.5) to the efficiency term to mitigate the dominance of ``speed chasing'' in the final score, ensuring that constraints such as safety and comfort retain sufficient discriminative power.

2) *Traffic Efficiency*: Traffic efficiency measures an algorithm's ability to complete the task in a given scenario in a timely manner and serves as the foundational component in closed-loop scenario evaluation. OVPD decomposes efficiency into a binary pass/fail base score and a continuous travel-time score. Passing the scenario yields a base score of 30 points, while the remaining 70 points are determined by the ratio between the expected travel time and the actual travel time:

$$G_{\mathrm{eff}} = \mathrm{pass} \times 30 + 70 \times \frac{t_E}{t_{\mathrm{pass}}} \quad (2)$$

where,

$\mathrm{pass} \in \{0,1\}$ indicates whether the scenario is successfully completed; if not, the efficiency score is set to $0$;

$t_{\mathrm{pass}}$ is the actual time for the ego vehicle to travel from the scenario trigger point to the termination point, **including any stopping/waiting time**;

$t_E$ is the expected travel time, defined as the time required to traverse the scenario at a constant speed of 25km/h. Since maintaining 25km/h without slowing down is rarely achievable in complex urban interactive scenarios, the inequality $\frac{t_E}{t_{\mathrm{pass}}} \le 1$ holds in practice, ensuring that $G_{\mathrm{eff}}$ strictly lies in $[0,100]$.

3) *Safety*: OVPD adopts a safety-floor principle: the safety dimension has a maximum of 10 points, and the occurrence of any safety-critical failure event sets this score to zero (hard zero), ensuring clear separability with respect to deployment risk. Safety failure events fall into three categories: (i)**Collision:** the ego vehicle physically collides with any traffic participant or road infrastructure; (ii)**Off-road:** the ego trajectory exhibits off-road behavior;(iii)**Takeover:** a safety-driver takeover in real-vehicle tests, serving as direct evidence of on-vehicle safety risk.

$$G_{safety} = \begin{cases} 10, \text{no collision, off-road or takeover} \\ 0, \text{otherwise} \end{cases} \quad (3)$$

4) *Comfort*: Comfort is closely tied to passenger experience and is primarily determined by longitudinal/lateral acceleration, jerk, and vehicle attitude changes (e.g., yaw rate) (De Winkel et al., 2023). OVPD assigns a maximum comfort score of 10 and applies a threshold-based scheme that penalizes the **fraction of time exceeding any comfort threshold** until the score is exhausted. Let $t_{\text{exceed}}$ denote the cumulative duration in which any comfort threshold is violated, and let $t_{\text{all}}$ denote the total scenario duration. We define:

$$G_{\text{comfort}} = 10 \times \left(1 - \frac{t_{\text{exceed}}}{t_{\text{all}}}\right) \tag{4}$$

The threshold set is specified by experts based on urban driving practice and is instantiated in implementation as a configuration file (details can be provided in the appendix).

5) *Traffic Coordination*: Traffic coordination measures how the ego vehicle affects the mobility of surrounding participants and the stability of local traffic flow (Ma et al., 2023; Yu et al., 2021). In OVPD, traffic coordination has a maximum of 10 points and is reported as $G_{\text{coord}} \in [0,10]$. We identify the set of affected background vehicles using a DBSCAN-like selection centered at the ego vehicle with an influence radius of $r = 30\,\text{m}$. For the vehicles within this set, we compute their composite scores and normalize it to the range [0,10] as follows:

$$G_{\text{coord}} = \frac{1}{N}\sum_{i=1}^{N}\left({G_{\text{eff}}^{i}}^{0.5} \times \left(0.6\,G_{\text{safety}}^{i} + 0.4\,G_{\text{comfort}}^{i}\right)\right) \tag{5}$$

Here, $G_{\text{eff}}^{i}$, $G_{\text{safety}}^{i}$, and $G_{\text{comfort}}^{i}$ are computed in the same way as defined in the preceding sections. We note that traffic coordination can be influenced by intrinsic properties of the traffic scenario itself; nevertheless, it remains informative for cross-method comparative diagnosis and provides a qualitative indicator of a single method's real-world interaction quality.

6) *Rule Compliance*: Compliant driving is a prerequisite for deploying autonomous vehicles on public roads (Liu et al., 2025; Manas and Paschke, 2023). Following common high-frequency violation types on urban roads in mainland China, OVPD performs event-level penalties over the following categories:

**Signal-related violations:** e.g., running a red light, entering prohibited segments or restricted lanes such as bus-only lanes;

**Yielding-related violations:** e.g., failing to yield to vehicles or pedestrians as required;

**Maneuver violations:** e.g., illegal U-turns, wrong-way driving, speeding, and illegal stopping/parking (as specified by scenario configurations and event-log definitions).

Rule compliance has a maximum of 10 points and uses an equal-weight, event-count penalty scheme: each violation event deducts 1 point until the score is exhausted:

$$G_{\text{rule}} = \max(0,\ 10 - n_{\text{violation}}) \tag{6}$$

where $n_{\text{violation}}$ is the number of violation events triggered in the scenario. The event log records each violation with its *type--location--timestamp*, enabling replay-based verification and error attribution.

### *3.4. Full-Scale Vehicle Deployment*

To narrow the gap in vehicle dynamics and execution stacks between purely simulation-based closed-loop testing and real-world deployment, OVPD provides an **optional Physical Testing enhancement module**. Built upon the Onsite virtual--physical fusion platform (VP-AutoTest), this module maps selected key atomic scenarios from a scenario chain to real-vehicle execution on a closed proving ground, thereby obtaining real vehicle dynamics feedback and evaluating an algorithm's stability and robustness under real-time constraints. The overall platform architecture---covering virtual traffic generation, perception-layer fusion, safety operation and takeover procedures, among other capabilities---is not repeated here; we instead refer to the system design and engineering implementation of VP-AutoTest (Cui et al., 2025).

1) *On-Vehicle Deployment and Closed-Loop Execution*

In OVPD physical testing, the evaluated algorithm can be deployed via a standardized engineering framework onto an L4-capable autonomous vehicle equipped with a multi-sensor suite and a drive-by-wire chassis. The on-vehicle middleware uses CyberRT (Baidu Apollo Team, 2019), through which the algorithm receives task and environment information via standardized communication interfaces and outputs control commands to execute the scenario in closed loop. VP-AutoTest provides a reference implementation pattern for the vehicle-side domain controller and the communication pipeline between the vehicle and the platform.

2) *Real-Time Constraints and Dynamics Feedback*:

Unlike simulation, which allows pausing and step-wise debugging, real-vehicle execution requires the algorithm to satisfy strict real-time control rates. OVPD enforces a minimum control output frequency of **10 Hz** in physical tests: the planning/decision module outputs control signals, which are then executed by the vehicle controller and drive-by-wire chassis in a low-level closed loop. This setup can effectively expose engineering issues that are often masked in simulation, such as control-output jitter, accumulated actuation latency, policy hysteresis, and stability degradation in highly interactive scenarios.

Physical tests provide authentic dynamics feedback, including vehicle speed, acceleration, steering response, and the true pose evolution during trajectory tracking. Developers can leverage these signals to tune the smoothness and stability of control outputs and to compare behavioral discrepancies between simulation and real vehicles under identical scenario semantics. VP-AutoTest further emphasizes establishing a unified reference frame via spatiotemporal synchronization and high-definition maps, supporting virtual--physical consistency checks and reliable evaluation.

Overall, OVPD Physical Testing serves as an optional enhancement module whose core value lies in introducing real vehicle dynamics and strict real-time constraints **without changing scenario semantics or the evaluation framework**. This enables complementary verification of critical risks---particularly cases where an algorithm ``passes in simulation but behaves unstably on-vehicle''---thereby improving the practical relevance of evaluation results for deployment.

### *3.5. Dataset formats*

With the above pipeline, the OVPD dataset can be constructed. The dataset is organized in a unified format, and all components can be replayed consistently using global timestamps:

**Ego control command and motion state:** The Ego vehicle's control signals, including throttle, brake, and steering wheel angle, as well as ego motion states, including global trajectory, velocity, heading (yaw), and angular acceleration, are recorded in JSON files at 10Hz.

**Surround-view and BEV renderings:** The test process is synchronously modeled in the digital-twin platform, which generates, temporal-ordered camera-rendered images for each data frame.

**Participants' trajectories and motion states:** For all agents in the background traffic flow, including vehicles, pedestrians, and cyclists, their agent category and motion states---including global trajectory, velocity, heading (yaw), and angular acceleration---are recorded in JSON files at 10~Hz.

**Traffic light states:** The signal states of each signal-controlled region in the map, together with its corresponding influence area, are logged to support rule constraints and compliance-event determination.

**HD map:** HD maps in OpenDRIVE format (Dupuis et al., 2010)

are provided for the two test sites, describing static elements in the proving grounds such as lanes, intersections, and traffic lights.

**Evaluation score:** The system automatically computes the overall driving score as well as per-dimension sub-scores for that scenario, enabling both per-scenario and overall benchmarking and diagnostic analysis.

**Table 2.** OVPD Baseline leaderboard (real-vehicle in-the-loop)

| Team | Overall | Safe | Eff | Comf | Comp | Coord | Pass |
|---|---|---|---|---|---|---|---|
| T11 | 89.44 | 100.00 | 87.39 | 94.18 | 96.00 | 81.34 | 1.00 |
| T17 | 89.42 | 100.00 | 84.58 | 97.81 | 100.00 | 85.24 | 1.00 |
| T14 | 89.29 | 100.00 | 85.10 | 96.69 | 100.00 | 85.13 | 1.00 |
| T16 | 88.38 | 100.00 | 83.87 | 96.19 | 99.33 | 83.99 | 1.00 |
| T6 | 85.61 | 100.00 | 84.59 | 88.22 | 89.67 | 84.67 | 1.00 |
| T15 | 83.98 | 93.33 | 80.30 | 91.85 | 92.67 | 77.00 | 0.93 |
| T20 | 82.17 | 93.33 | 80.48 | 85.14 | 92.00 | 77.37 | 0.93 |
| T5 | 81.79 | 100.00 | 73.03 | 93.97 | 97.67 | 86.72 | 1.00 |
| T7 | 80.39 | 100.00 | 69.97 | 97.45 | 96.00 | 80.23 | 1.00 |
| T8 | 79.37 | 100.00 | 69.48 | 95.71 | 94.33 | 79.27 | 1.00 |
| T9 | 78.96 | 100.00 | 66.78 | 96.41 | 95.33 | 87.29 | 1.00 |
| T1 | 75.53 | 93.33 | 64.33 | 91.94 | 88.67 | 83.26 | 0.93 |
| T3 | 75.44 | 93.33 | 65.98 | 90.79 | 88.67 | 74.30 | 0.93 |
| T4 | 70.86 | 86.67 | 61.96 | 84.42 | 82.67 | 75.46 | 0.87 |
| T18 | 69.00 | 86.67 | 58.63 | 83.59 | 86.00 | 71.64 | 0.87 |
| T2 | 61.39 | 80.00 | 52.08 | 73.72 | 79.33 | 64.79 | 0.80 |
| T21 | 61.33 | 80.00 | 49.10 | 79.31 | 79.33 | 68.52 | 0.80 |
| T10 | 57.41 | 66.67 | 53.23 | 64.03 | 65.33 | 54.01 | 0.67 |
| T12 | 49.88 | 66.67 | 38.89 | 65.32 | 66.67 | 58.25 | 0.67 |
| T19 | 29.60 | 40.00 | 23.48 | 38.71 | 40.00 | 31.99 | 0.40 |
| avg | 73.96 | 89.00 | 66.66 | 85.27 | 86.48 | 74.52 | 0.89 |

## 4 Statistics and Evaluation

In this section, we summarize the distributional metrics of OVPD from a dataset perspective, focusing on its coverage, difficulty distribution, discriminability, and diagnostic granularity, rather than ranking individual methods.

All replay clips in the dataset are competition records of participating algorithms from the 2025 Onsite Challenge, collected under a vehicle-in-the-loop testing setup and across the scenarios described in Table 1. Scores are computed by the metrics in Sec 3.3. and listed in Table 2., which provide an overall view of the dataset's difficulty distribution across scenarios and capability dimensions, as well as the discriminability of individual scenarios:

***1) Category-level trends reflect dataset coverage and difficulty gradients:*** The category-level data distribution indicates that OVPD provides structured coverage across emergency response, traffic efficiency, and rule compliance, instead of concentrating on a single driving competency. This supports its use as a capability-oriented dataset. According to Table 3. and Fig. 3, Category A includes more hard-failure characteristics, Category B places greater pressure on interaction efficiency, and Category C imposes the strongest compliance constraints, reflected by the clips' lowest pass rate and broadest metric degradation.

**Table 3.** Category-level results on OVPD baseline

| Cat. | Overall | Pass | Safety | Eff. | Comf. | Compl. | Coord. |
|---|---|---|---|---|---|---|---|
| **A** | 78.05 | 0.91 | 90.71 | 72.73 | 86.29 | 88.75 | 76.46 |
| **B** | 73.81 | 0.93 | 92.50 | 64.09 | 88.26 | 89.88 | 76.05 |
| **C** | 66.95 | 0.83 | 82.50 | 58.61 | 80.52 | 79.12 | 69.62 |

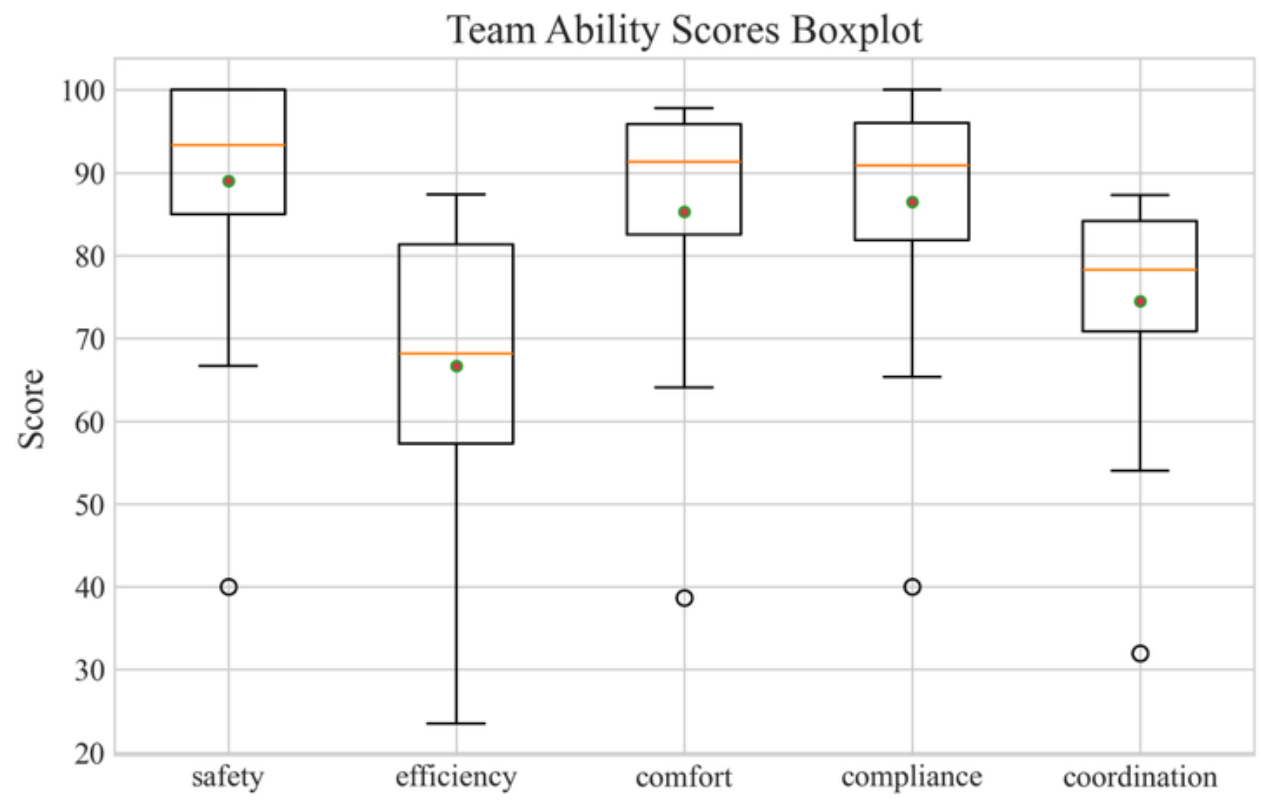

**Fig. 4.** Box plot of the scores of 20 teams calculated based on their ability scores

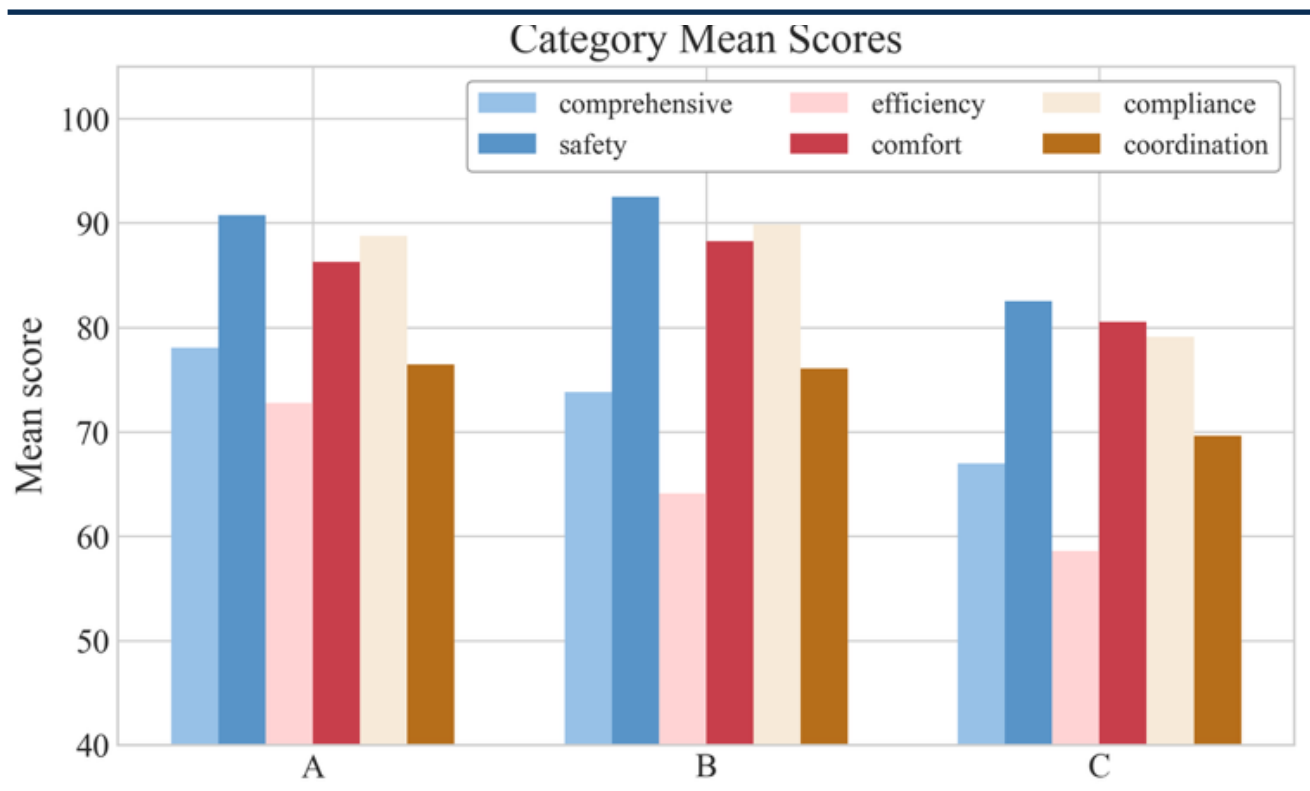

**Fig. 3.** The average values of all teams' various ability scores in all atomic scenarios are calculated based on the category of the scenario

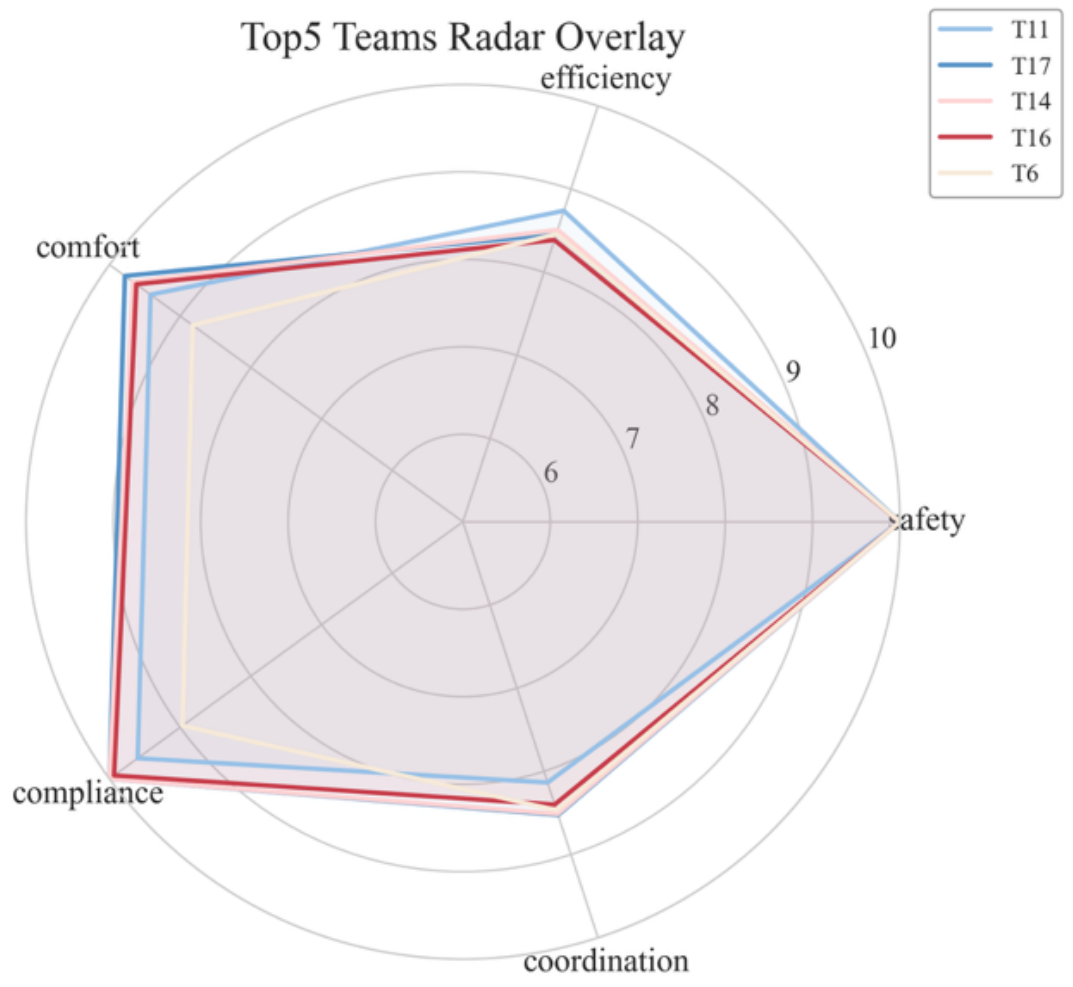

**Fig. 5.** The performance scores of the top 5 teams in the Overall Leaderboard across 5 different categories

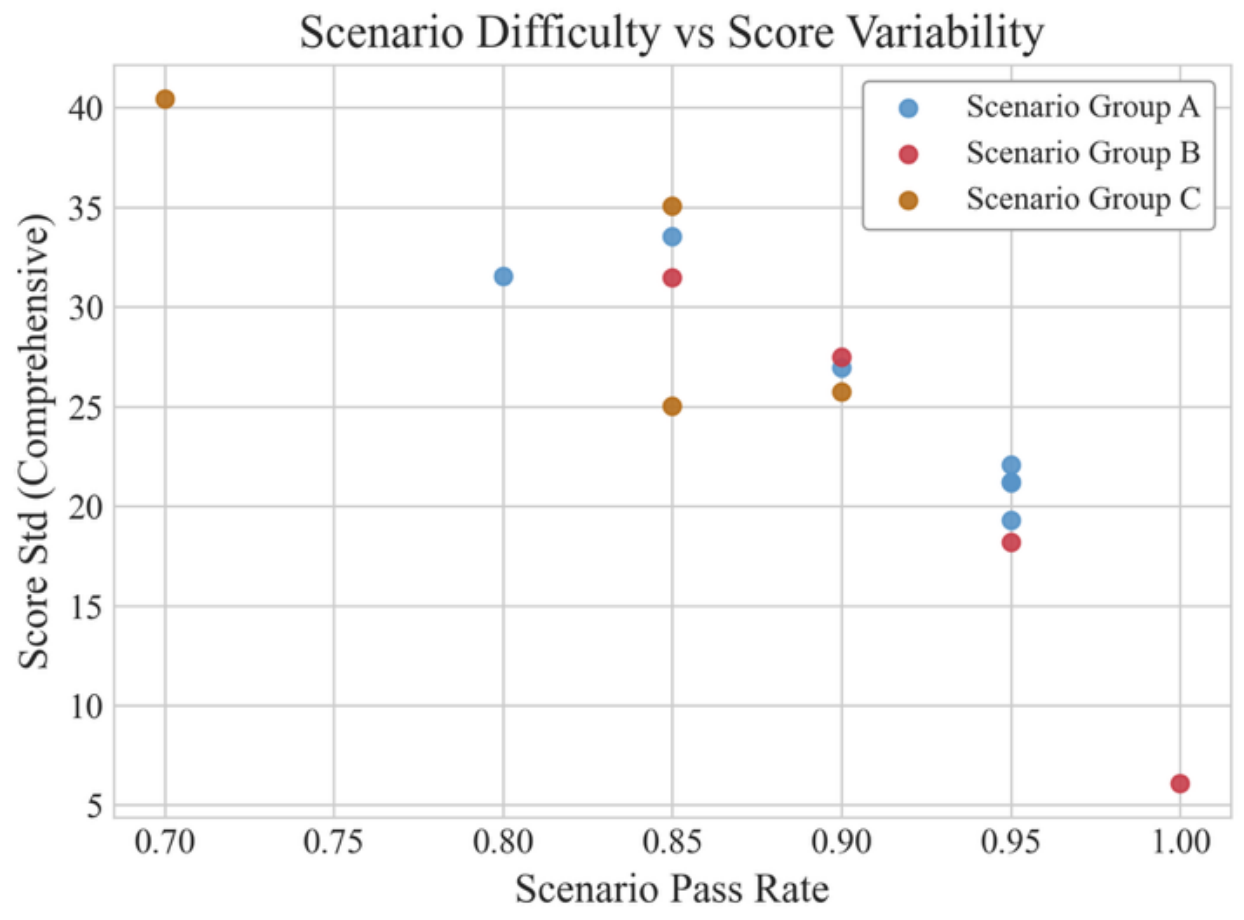

**Fig. 6.** A two-dimensional scatter plot showing the variance of scores for each team in every atomic scenario and the team pass rate. Each dot in the figure represents an atomic scene, and a different color represents a different scene category

**2) Metrics reveal different performance distributions across dimensions:** The multi-dimensional metrics show that OVPD contains distinct performance distributions across different evaluation dimensions. As shown in Fig. 4, algorithm results are more dispersed in efficiency and traffic coordination, while remaining relatively more concentrated in other dimensions. Fig. 5 further shows that algorithms with similar overall scores can still present different profiles across individual dimensions. This indicates that OVPD preserves heterogeneous behavior distributions across safety, efficiency, comfort, rule compliance, and coordination, rather than collapsing them into a single aggregate distribution.

**3) Scenario diversity supports layered dataset structure:**Within the 15-scenario Onsite baseline, OVPD includes both foundational scenarios and more discriminative ones. As shown in Fig. 6, some scenarios present relatively concentrated metric distributions, reflecting basic capability checks with clear task requirements, while others show broader variation by more complex interactions. This diversity gives OVPD a layered structure that supports essential coverage, richer behavioral analysis, and future extension.

## 5 Conclusion

This paper releases OVPD (OnSite Virtual--Physical Dataset), a virtual--physical fusion testing dataset derived from the 2025 Onsite Autonomous Driving Challenge, providing deployment-oriented data support for training, replay-based analysis, and diagnosis of planning/decision-making algorithms in complex interactive scenarios. Centered on a real ego vehicle in the loop, OVPD integrates simulated background traffic and vehicle--roadside perception within a closed-course test site, yielding a controllable and interactive testing environment as well as data collection with real vehicle dynamics feedback. The dataset covers complete test runs from 20 teams over a scenario chain consisting of 15 atomic scenarios, totaling 20 clips (nearly 3 hours). All modalities are aligned by a unified global timestamp, including ego trajectories and motion states, control commands, and digital-twin-rendered surround-view and BEV observations. In addition, OVPD releases five-dimensional score results together with the corresponding computation protocols, and aggregates performance at the single-scenario granularity, providing a consistent and reproducible basis for cross-method comparison and scenario-level capability profiling.

## 6 Future Works

Looking ahead, there are several important directions for extending OVPD. First, although the current platform already supports deployment-oriented closed-loop testing through a *real-vehicle-in-the-loop + simulated background traffic + digital-twin rendering* pipeline, future iterations can further improve the realism of interactive background traffic. At present, surrounding traffic participants are mainly generated by the microscopic traffic simulator TESSNG (Jida Traffic, 2026). A promising next step is to incorporate richer real-world driving priors and more socially aware or adversarial behavior models, so as to better capture the diversity and subtlety of multi-agent interactions in real traffic. This would help narrow the gap between the interaction complexity of OVPD and that of real-world deployment, especially in highly interactive urban scenarios.

Second, future development of OVPD will continue to expand scenario breadth and diversity while preserving its capability-oriented diagnostic design. The current OVPD data collection and Onsite scenario library construction mainly focus on targeted coverage of challenging urban capability deficiencies. Building on this foundation, future work may combine real-world datasets and operational driving logs to mine high-value segments under a capability-oriented principle, and further parameterize them into atomic scenarios. Such an extension would progressively enrich scenario diversity and coverage, while maintaining the dataset's utility for targeted testing and capability diagnosis.

## Author contributions

**Yuhang Zhang:** Writing – original draft, Validation, Software, Methodology, Formal analysis, Data curation, Conceptualization.

**Jiarui Zhang:** Concept Designation, Review, Supervision, Methodology.

**Bowen Jian:** Concept Designation, Review, Supervision, Methodology.

**Xin Zhou:** Concept Designation, Review, Supervision, Methodology.

**Zhichao Lv:** Review & editing, Formal Analysis, Methodology.

**Peng Hang:** Review & editing, Formal Analysis, Methodology.

**Rongjie Yu:** Review & editing, Formal Analysis, Methodology.

**Ye Tian:** Review & editing, Formal Analysis, Methodology.

**Jian Sun:** Review & editing, Formal Analysis, Methodology.

## Replication and data sharing

For reproducibility and verification, we have organized and open-sourced the main data. The raw data can be found at:

https://huggingface.co/datasets/Yuhang253820/Onsite_OPVD

## Acknowledgements

This work was supported in part by the National Natural Science Foundation of China (52125208, 52232015) and the Fundamental and Interdisciplinary Disciplines Breakthrough Plan of the Ministry of Education of China (No.JYB2025XDXM123).

## Declaration of competing interest

The authors have no competing interests to declare that are relevant to the content of this article.

## Graphical abstract and Highlights

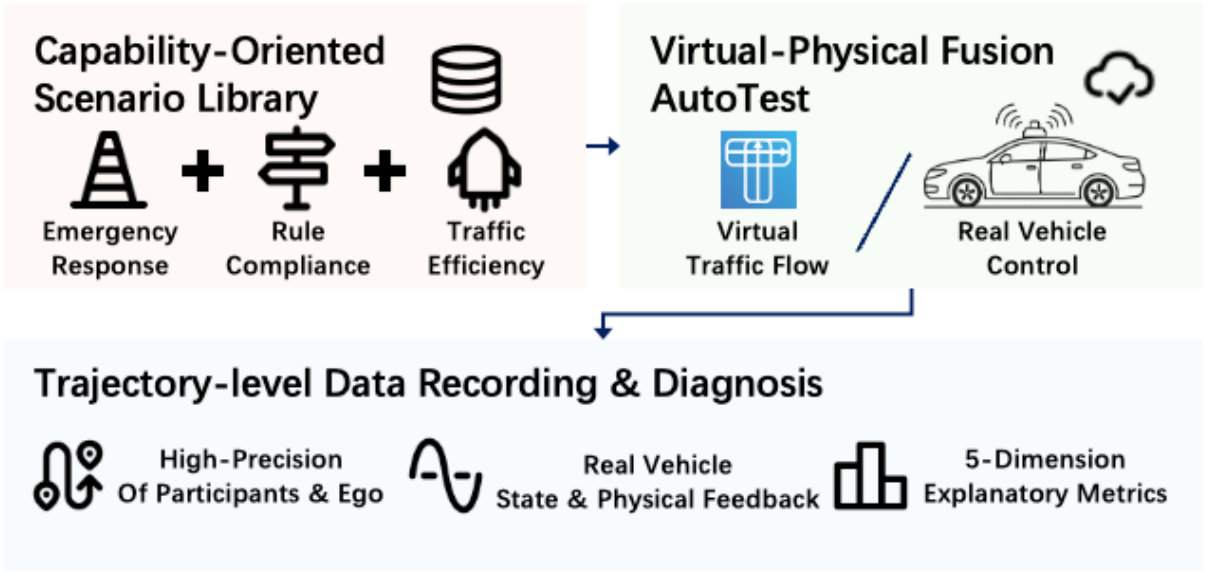

**Graphical abstract of OVPD**

**Highlights:**

- Deployment-oriented virtual–physical fusion dataset for autonomous driving. We present OVPD, a virtual–physical fusion testing dataset derived from the 2025 Onsite Autonomous Driving Challenge, which integrates virtual traffic with reality perception to construct controllable and interactive closed-loop evaluation environments.
- Capability-driven scenario construction with controllable difficulty. OVPD adopts a hierarchical scenario design across driving functions, road facilities, and core capabilities, enabling targeted long-tail coverage and fine-grained difficulty adjustment for diagnostic testing.
- Scenario-centric evaluation with multi-dimensional metrics. The dataset introduces a unified evaluation framework across safety, efficiency, comfort, rule compliance, and traffic coordination, supporting scenario-level capability diagnosis and interpretable benchmarking.
- Real vehicle dynamics and closed-loop interaction data. OVPD captures authentic vehicle dynamics, control commands, and execution feedback under real-time constraints, bridging the gap between simulation-based evaluation and real-world deployment.
- Comprehensive multi-modal data for training and analysis. The dataset provides synchronized trajectories, control signals, digital-twin-rendered views, and evaluation scores, supporting open-loop validation, closed-loop testing, and failure diagnosis.

## References

Baidu Apollo Team, 2019. Apollo: An open-source runtime framework for autonomous driving. GitHub. Accessed: 2024-03-19.

Caesar, H., Bankiti, V., Lang, A. H., Vora, S., Liong, V. E., Xu, Q., et al., 2020. nuScenes: A multimodal dataset for autonomous driving. Proceedings of the IEEE/CVF Conference on Computer Vision and Pattern Recognition, pp. 11621–11631.

Crosato, L., Tian, K., Shum, H. P. H., Ho, E. S. L., Wang, Y., Wei, C., 2023. Social interaction-aware dynamical models and decision making for autonomous vehicles.

Cui, Y., Fang, S., Zhang, J., Huang, Y., Xu, C., Zhu, B., et al., 2025. VP-AutoTest: A virtual-physical fusion autonomous driving testing platform. arXiv preprint arXiv:2512.07507.

Daza, I. G., Izquierdo, R., Martinez, L. M., Benderius, O., Llorca, D. F., 2023. Sim-to-real transfer and reality gap modeling in model predictive control for autonomous driving. Applied Intelligence, 53(10), 12719–12735.

De Winkel, K. N., Irmak, T., Happee, R., Shyrokau, B., 2023. Standards for passenger comfort in automated vehicles: Acceleration and jerk. Applied Ergonomics, 106, 103881.

Dosovitskiy, A., Ros, G., Codevilla, F., Lopez, A., Koltun, V., 2017. CARLA: An open urban driving simulator. Proceedings of the 1st Annual Conference on Robot Learning, pp. 1–16.

Dupuis, M., Strobl, M., Grezlikowski, H., 2010. OpenDRIVE 2010 and beyond: Status and future of the de facto standard for the description of road networks. Proceedings of the Driving Simulation Conference Europe 2010, pp. 231–242.

Ericsson, E., 2000. Variability in urban driving patterns. Transportation Research Part D: Transport and Environment, 5(5), 337–354.

Ettinger, S., Cheng, S., Caine, B., Liu, C., Zhao, H., Pradhan, S., et al., 2021. Large scale interactive motion forecasting for autonomous driving: The Waymo Open Motion Dataset.

Fan, H., Zhu, F., Liu, C., Zhang, L., Zhuang, L., Li, D., et al., 2018. Baidu Apollo EM motion planner. arXiv preprint arXiv:1807.08048.

Gao, S., Yang, J., Chen, L., Chitta, K., Qiu, Y., Geiger, A., et al., 2024. Vista: A generalizable driving world model with high fidelity and versatile controllability. Advances in Neural Information Processing Systems.

Garcia Daza, I., Izquierdo, R., Martinez, L. M., Benderius, O., Fernandez Llorca, D., 2023. Sim-to-real transfer and reality gap modeling in model predictive control for autonomous driving. Applied Intelligence, 53, 12719–12735.

Guo, Y., Xu, C., Liu, J., Zhang, H., Hang, P., Sun, J., 2025. Interactive adversarial testing of autonomous vehicles with adjustable confrontation intensity. arXiv preprint arXiv:2507.21814.

Hafner, D., Pasukonis, J., Ba, J., Lillicrap, T., 2025. Mastering diverse control tasks through world models. Nature, pp. 1–7.

Jia, X., Yang, Z., Li, Q., Zhang, Z., Yan, J., 2024. Bench2Drive: Towards multi-ability benchmarking of closed-loop end-to-end autonomous driving. Advances in Neural Information Processing Systems, 37, 819–844.

Jiang, X., Zhao, X., Liu, Y., Li, Z., Hang, P., Xiong, L., et al., 2025. A naturalistic trajectory dataset with dense interaction for autonomous driving. Scientific Data, 12, 1084.

Jida Traffic, 2026. TESSNG: Traffic simulation software (official website). Accessed: 2026-02-26.

Karnchanachari, N., Geromichalos, D., Tan, K. S., Li, N., Eriksen, C., Yaghoubi, S., et al., 2024. Towards learning-based planning: The nuPlan benchmark for real-world autonomous driving. 2024 IEEE International Conference on Robotics and Automation, pp. 629–636.

Kerbl, B., Kopanas, G., Leimkuhler, T., Drettakis, G., 2023. 3D Gaussian splatting for real-time radiance field rendering. ACM Transactions on Graphics, 42(4).

Kurenkov, M., Marvi, S., Schmidt, J., Rist, C. B., Canevaro, A., Yu, H., et al., 2024. Traffic and safety rule compliance of humans in diverse driving situations. arXiv preprint.

Kurenkov, M., Marvi, S., Schmidt, J., Rist, C. B., Canevaro, A., Yu, H., et al., 2024. Traffic and safety rule compliance of humans in diverse driving situations. arXiv preprint.

Li, Z., Yu, Z., Lan, S., Li, J., Kautz, J., Lu, T., et al., 2024. Is ego status all you need for open-loop end-to-end autonomous driving? Proceedings of the IEEE/CVF Conference on Computer Vision and Pattern Recognition, pp. 14864–14873.

Liu, H., Chen, K., Li, Y., Huang, Z., Liu, M., Ma, J., 2025. UDMC: Unified decision-making and control framework for urban autonomous driving with motion prediction of traffic participants. IEEE Transactions on Intelligent Transportation Systems.

Lou, G., Deng, Y., Zheng, X., Zhang, M., Zhang, T., 2022. Testing of autonomous driving systems: Where are we and where should we go? Proceedings of the 30th ACM Joint European Software Engineering Conference and Symposium on the Foundations of Software Engineering, pp. 31–43.

Ma, L., Qu, S., Ren, J., Zhang, X., 2023. Mixed traffic flow of human-driven vehicles and connected autonomous vehicles: String stability and fundamental diagram. Mathematical Biosciences and Engineering, 20(2), 2280–2295.

Manas, K., Paschke, A., 2023. Legal compliance checking of autonomous driving with formalized traffic rule exceptions. ICLP Workshops.

Monroe, W. S., 1923. Introduction to the theory of educational measurement. Houghton Mifflin, Boston, MA.

National Transportation Safety Board, 2019. Collision between vehicle controlled by developmental automated driving system and pedestrian, Tempe, Arizona, March 18, 2018. Highway Accident Report NTSB/HAR-19/03.

OnSite Autonomous Driving Public Testing Service Platform, n.d. Onsite autonomous driving connected joint testing real-vehicle competition: Participant handbook. Online handbook page (Chinese).

Research Team for Traffic OPerations and Simulation, 2025. The 2025 Onsite autonomous driving connected joint testing real-vehicle competition concludes successfully, marking new breakthroughs in autonomous driving evaluation research.

Sadigh, D., Sastry, S., Seshia, S. A., Dragan, A. D., 2016. Planning for autonomous cars that leverage effects on human actions. Robotics: Science and Systems, 2, 1–9.

SAE International, 2021. Taxonomy and definitions for terms related to driving automation systems for on-road motor vehicles. Standard J3016 202104, SAE International, Warrendale, PA.

Stocco, A., Pulfer, B., Tonella, P., 2022. Mind the gap! A study on the transferability of virtual versus physical-world testing of autonomous driving systems. IEEE Transactions on Software Engineering, 49(4), 1928–1940.

Stocco, A., Pulfer, B., Tonella, P., 2023. Mind the gap! A study on the transferability of virtual versus physical-world testing of autonomous driving systems. IEEE Transactions on Software Engineering, 49, 1928–1940.

Sun, H., Feng, S., Yan, X., Liu, H. X., 2021. Corner case generation and analysis for safety assessment of autonomous vehicles. Transportation Research Record, 2675(11), 587–600.

Sun, P., Kretzschmar, H., Dotiwalla, X., Chouard, A., Patnaik, V., Tsui, P., et al., 2020. Scalability in perception for autonomous driving: Waymo Open Dataset. Proceedings of the IEEE/CVF Conference on Computer Vision and Pattern Recognition, pp. 2446–2454.

United Nations Economic Commission for Europe, 2025. Innovation and application of testing methodology on traffic rule compliance for ADS-equipped vehicles. Informal document GRVA-23-44e, UNECE.

Wang, P., Huang, X., Cheng, X., Zhou, D., Geng, Q., Yang, R., 2019. The ApolloScape open dataset for autonomous driving and its application. IEEE Transactions on Pattern Analysis and Machine Intelligence, 1.

Xu, R., Lin, H., Jeon, W., Feng, H., Zou, Y., Sun, L., et al., 2025. WOD-E2E: Waymo Open Dataset for end-to-end driving in challenging long-tail scenarios.

Yang, S., Wang, C., Zhang, Y., Yin, Y., Huang, Y., Li, S. E., et al., 2024.

Quantitative representation of scenario difficulty for autonomous driving based on adversarial policy search. Research, 8.

Yan, Y., Lin, H., Zhou, C., Wang, W., Sun, H., Zhan, K., et al., 2024. Street Gaussians: Modeling dynamic urban scenes with Gaussian splatting. Computer Vision – ECCV 2024. Springer.

Yu, H., Jiang, R., He, Z., Zheng, Z., Li, L., Liu, R., et al., 2021. Automated vehicle-involved traffic flow studies: A survey of assumptions, models, speculations, and perspectives. Transportation Research Part C: Emerging Technologies, 127, 103101.

Zhai, J.-T., Feng, Z., Du, J., Mao, Y., Liu, J.-J., Tan, Z., et al., 2023. Rethinking the open-loop evaluation of end-to-end autonomous driving in nuScenes. arXiv preprint arXiv:2305.10430.

Zhang, S., Chen, Q., Zhang, X., Qiu, J., Li, X., Li, Y., et al., 2025. Simulation study of freeway work zone scenes for autonomous driving using Simulink and PreScan. Scientific Reports, 15(1), 10058.

Zhao, G., Ni, C., Wang, X., Zhu, Z., Zhang, X., Wang, Y., et al., 2025. DriveDreamer4D: World models are effective data machines for 4D driving scene representation. Proceedings of the IEEE/CVF Conference on Computer Vision and Pattern Recognition.

Zhou, J., Wang, L., Meng, Q., Wang, X., 2025. Intelligence evaluation methods for autonomous vehicles. 2025 IEEE International Conference on Robotics and Automation, pp. 10600–10606.

Zhou, H., Liu, H., Lu, H., Ma, J., Ji, Y., 2024. Enhance planning with physics-informed safety controller for end-to-end autonomous driving. 2024 IEEE International Conference on Robotics and Biomimetics, pp. 1775–1782.

## *Author biography*

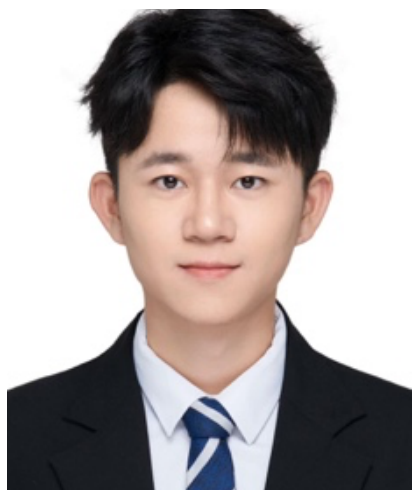

**Yuhang Zhang** received the B.S. degree in College of Transportation from Tongji University, Shanghai, China. He is currently pursuing the M.S. degree with the Department of Traffic Engineering, Tongji University, Shanghai, China. His main research interests include end-to-end autonomous driving algorithm, world model and reinforcement learning.

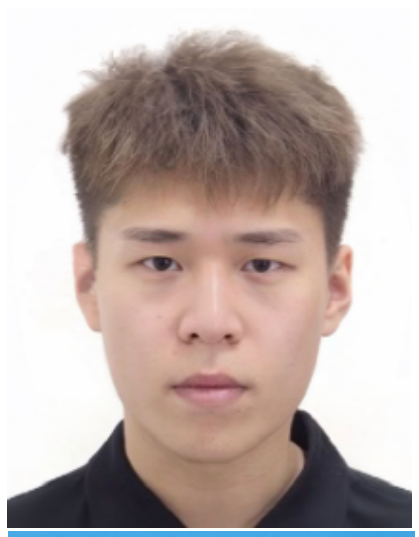

**Jiarui Zhang** received the B.S. degree in College of Civil Aviation from Nanjing University of Aeronautics and Astronautics, Nanjing, China. He is currently pursuing the Ph.D. degree with the Department of Traffic Engineering, Tongji University, Shanghai, China. His main research interests include autonomous driving testing and evaluation, and testing scenario generation.

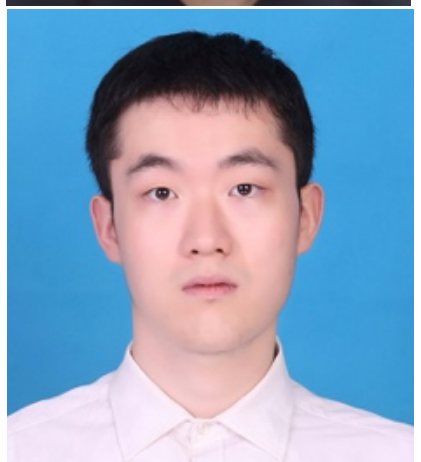

**Bowen Jian** received the B.S. degree in College of Transportation from Tongji University, Shanghai, China, in 2023. He is currently pursuing the Ph.D. degree with the Department of Traffic Engineering, Tongji University, Shanghai, China. His main research interests include law-compliance testing and enhancement for autonomous vehicles.

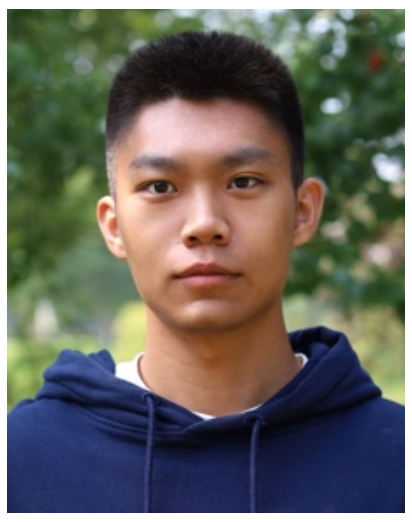

**Xin Zhou** received the B.S. degree in transportation engineering from Tongji University, Shanghai, China. He is currently pursuing the Ph.D. degree with the Department of Traffic Engineering, Tongji University, Shanghai, China. His research interests include fault diagnosis and regulation for highly automated vehicles.

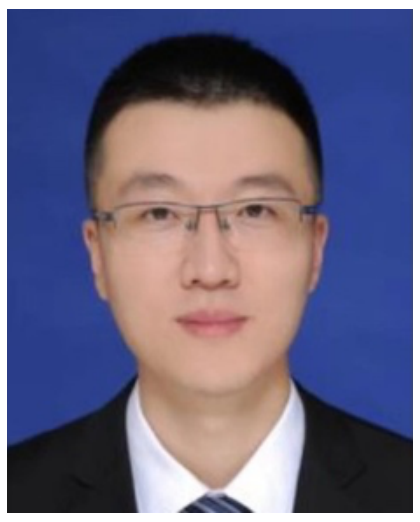

**Zhichao Lv** is a Postdoctoral Researcher with the College of Transportation Engineering, Tongji University. He received the B.Eng. degree from Taiyuan University of Technology in 2010 and the M.Sc. and Ph.D. degrees in control engineering and vehicle engineering from Jilin University and Tongji University in 2018 and 2023, respectively. His research interests include human-like decision-making and reinforcement learning for intelligent driving, and high-performance in-vehicle computing platforms.

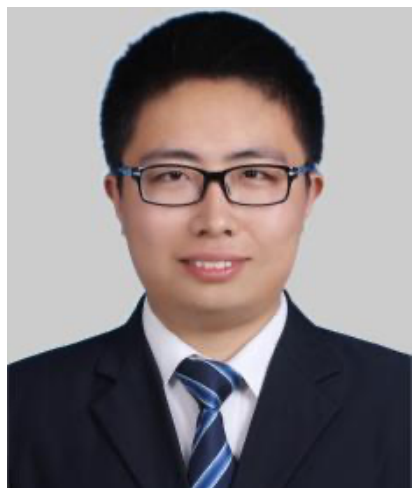

**Peng Hang** is an Associate Professor at the Department of Traffic Engineering, Tongji University, Shanghai, China. He received the Ph.D. degree with the School of Automotive Studies, Tongji University, Shanghai, China, in 2019. He was a Visiting Researcher with the Department of Electrical and Computer Engineering, National University of Singapore, Singapore, in 2018. From 2020 to 2022, he served as a Research Fellow with the School of Mechanical and Aerospace Engineering, Nanyang Technological University, Singapore. His research interests include vehicle dynamics and control, decision making, motion planning and motion control for autonomous vehicles. He serves as an Associate Editor of IEEE Internet of Things Journal, IEEE Transactions on Vehicular Technology, Journal of Field Robotics, IET Smart Cities, and SAE International Journal of Vehicle Dynamics, Stability, and NVH.

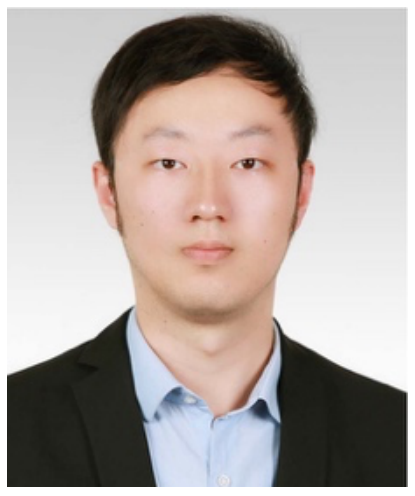

**Rongjie Yu** received the B.Sc. degree from TongjiUniversity in 2010 and the M.Sc. and Ph.D. degreesin traffic engineering from the University of Central Florida in 2012 and 2013, respectively. He iscurrently an Associate Professor with the Collegeof Transportation Engineering, Tongji University.His research interests include traffic safety, humanbehavior, and safety evaluation of connected andautonomous vehicles.

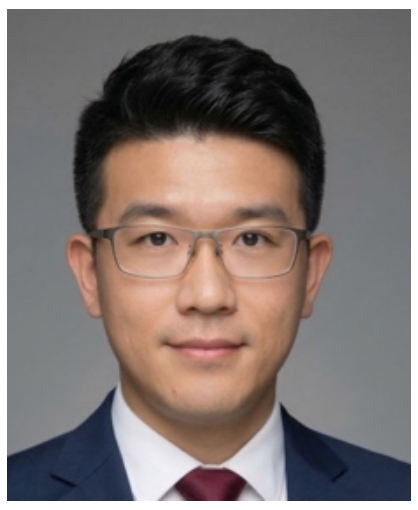

**Ye Tian** received the Ph.D. degreein transportation engineering from The University ofArizona, Tucson, AZ, USA, in 2015. He is currentlyan Associate Professor of transportation engineering with Tongji University, Shanghai, China. Hisresearch interests include active demand manage-ment, dynamic traffic assignment, mesoscopic traf-fic simulation, and safety assurance of automatedvehicles.

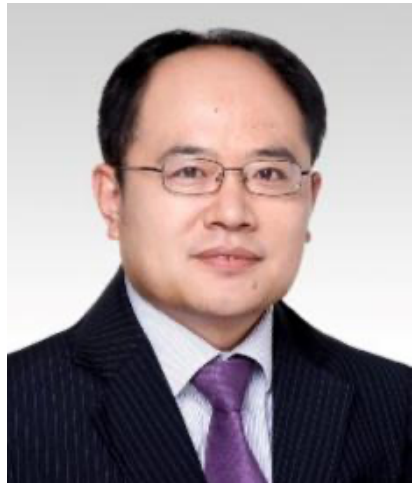

**Jian Sun** received the Ph.D. degree from Tongji University in 2006. Subsequently, he was at Tongji University as a Lecturer, and then promoted to the position as a Professor in 2011, where he is currently a Professor with the College of Transportation Engineering and the Dean of the Department of Traffic Engineering. His main research interests include traffic flow theory, traffic simulation, connected vehicleinfrastructure system, and intelligent transportation system.